\documentclass[12pt]{article}
\usepackage{graphicx, url}
\usepackage{palatino}
\usepackage{amsmath, amssymb,bbm}
\usepackage{colortbl}
\usepackage{caption}
\usepackage[table]{xcolor}
\usepackage{titlesec}
\usepackage{longtable}
\usepackage{xcolor}
\usepackage{booktabs}
\usepackage{multirow}
\usepackage{fancybox, graphicx}
\usepackage{epsfig}
\usepackage{rotating}
\usepackage{dcolumn}
\usepackage{bm}

\titleformat{\paragraph}
{\normalfont\normalsize\bfseries}{\theparagraph}{1em}{}
\titlespacing*{\paragraph}
{0pt}{3.25ex plus 1ex minus .2ex}{1.5ex plus .2ex}
\definecolor{darkgreen}{rgb}{0.09766 0.4375 0.09766}
\definecolor{darkblue}{rgb}{0.09766 0.09766 0.4375}
\newcommand{\comment}[1]{}
\begin{document}   
\baselineskip 18pt
\title{Integrating Boosted learning with Differential Evolution (DE) Optimizer:A Prediction of Groundwater Quality Risk Assessment in Odisha}
\author{
  Sonalika Subudhi $^1$\footnote{E-mail: sonalikasubudhi2002@gmail.com },~  
   Alok Kumar Pati$^1$\footnote{E-mail: alokpati6065@gmail.com, ~alokpati@soa.ac.in ~(Corresponding Author)},~ 
    Sephali Bose$^1$\footnote{E-mail: sephalibose2003@gmail.com},\\ 
   Subhasmita Sahoo$^1$\footnote{E-mail: sahoosubhasmita316@gmail.com}, ~
    Avipsa Pattanaik$^1$\footnote{E-mail: pattanaikavipsa17@gmail.com}, 
~and   Biswa Mohan Acharya$^1$\footnote{E-mail: biswaacharya@soa.ac.in} \\ [0.2cm]
    {\it \small 1.  Department of Computer Application, Siksha 'O' Anusandhan Deemed to be University,} \\ [-0.2cm]
    {\it \small Bhubaneswar,751030, Odisha, India.
}\\ [-2cm]
    {\it \small } \\
    {\it \small}\\
}
\date{}
\maketitle
\begin{abstract}
\noindent Groundwater is eventually undermined by human exercises, such as fast industrialization, urbanization, over-extraction, 
and contamination from agrarian and urban sources. From among the different contaminants, 
the presence of heavy metals like cadmium (Cd), chromium (Cr), arsenic (As), and lead (Pb) 
proves to have serious dangers when present in huge concentrations in groundwater. 
Long-term usage of these poisonous components may lead to neurological disorders, 
kidney failure and different sorts of cancer. To address these issues, this study 
developed a machine learning-based predictive model to evaluate the Groundwater Quality Index (GWQI) 
and identify the main contaminants which are affecting the water quality. 
It has been achieved with the help of a hybrid machine learning model i.e. LCBoost Fusion . The model has undergone several processes 
like data preprocessing, hyperparameter tuning using Differential Evolution (DE) optimization, 
and evaluation through cross-validation. The LCBoost Fusion model outperforms individual models 
(CatBoost and LightGBM), by achieving low RMSE (0.6829), MSE (0.5102), MAE (0.3147) and a high R$^2$ 
score of 0.9809. Feature importance analysis highlights Potassium (K), Fluoride (F) 
and Total Hardness (TH) as the most influential indicators of groundwater contamination. 
This research successfully demonstrates the application of machine learning in assessing 
groundwater quality risks in Odisha. The proposed LCBoost Fusion model offers a reliable and efficient 
approach for real-time groundwater monitoring and risk mitigation. These findings 
will help the environmental organizations and the policy makers to map out targeted 
places for sustainable groundwater management. Future work will focus on using remote sensing 
data and developing an interactive decision-making system for groundwater quality assessment.
\newcommand{\keyword}[1]{\textbf{#1}}

\end{abstract}
\textbf{keywords:} Groundwater Quality, LCBoost Fusion , Machine Learning, Differential Evolution, ~Feature Importance.

\maketitle

\section*{Highlights:}
\begin{itemize}
\item[1] LCBoost Fusion, a hybrid machine learning model combining LightGBM and CatBoost, optimized using Differential Evolution (DE) for accurate Groundwater Quality Index (GWQI) prediction.
\item[2] The model outperforms individual models, achieving superior predictive accuracy with RMSE of 0.6829, MSE of 0.5102, MAE of 0.3147, and an R² score of 0.9809.
\item[3] Feature importance analysis identifies Potassium (K), Fluoride (F), and Total Hardness (TH) as the most influential contaminants affecting groundwater quality, aiding targeted water management effort.
\end{itemize}

\section{Introduction}
Groundwater quality evaluation is crucial because groundwater is one of the most important sources of drinking water for millions of people \cite{1}. Many regions, especially rural and urban areas, depend on groundwater for domestic, agricultural, and industrial purposes \cite{2,3}. However, groundwater quality can be significantly impacted by both natural factors and human activities, leading to potential health risks and environmental issues \cite{3,4}. Therefore, assessing its chemical composition and identifying its pollutants is essential for ensuring its safety and sustainable use \cite{5}. The quality of groundwater is affected by both chemical and biological factors \cite{6}. Naturally occurring elements such as chloride, sodium, fluoride, and heavy metals can lead to groundwater pollution, making the water unsafe for consumption \cite{7,8,9}. Additionally, variations in soil composition, water table levels, and climate conditions can impact groundwater quality \cite{10,11}. Human activities also contribute significantly to groundwater pollution. Agricultural practices, including excessive fertilizer and pesticide use, lead to the infiltration of nitrate and phosphate compounds, which degrade water quality \cite{10,12}. Industrial waste disposal, mining operations, and leakage from landfills introduce harmful chemicals and heavy metals, contaminating groundwater sources \cite{7,13}. Furthermore, untreated sewage, improper sanitation systems, and wastewater seepage contribute to bacterial and viral contamination, increasing the risk of waterborne diseases \cite{12,14,15}. Evaluating groundwater quality is essential for protecting public health, sustaining agricultural productivity, and preserving ecosystems \cite{9,10}. With increasing industrialization and agricultural expansion, implementing continuous monitoring, pollution control measures, and sustainable groundwater management is critical to ensuring a safe and reliable water supply for future generations \cite{8,16}.

\par
Groundwater quality evaluation is essential for ensuring safe drinking water, sustainable agriculture, industrial productivity, and environmental conservation \cite{17}. As a primary water source for millions, groundwater must be free from harmful contaminants, yet it is often polluted by industrial waste, agricultural runoff, and improper disposal of chemicals \cite{18,19}. Contaminants such as heavy metals, nitrates, and microbial pathogens can pose severe health risks, leading to neurological disorders, kidney issues, and waterborne diseases \cite{20,21}. Regular assessment helps in the early detection of pollutants, allowing for timely intervention to protect public health \cite{22}. In agriculture, groundwater plays a vital role in irrigation, and poor-quality water with high salinity or chemical pollutants can degrade soil fertility, reducing crop yields \cite{23,24,25}. Industries also rely on groundwater for various processes, and contamination can lead to equipment damage, higher maintenance costs, and non-compliance with environmental regulations \cite{26,27,28}. Furthermore, polluted groundwater threatens ecosystems as toxins seep into lakes, rivers, and wetlands, harming aquatic life and biodiversity \cite{29,30}. Sustainable management of groundwater resources is crucial to prevent over-extraction and long-term depletion \cite{31,32}. Advanced technologies like remote sensing, GIS mapping, and machine learning models enhance groundwater monitoring, making contamination detection more efficient \cite{33,34}. Collaborative efforts between governments, researchers, and communities are necessary to implement effective conservation strategies \cite{35}. By prioritizing groundwater quality evaluation, societies can ensure a clean and reliable water supply for future generations, safeguarding public health, economic stability, and environmental sustainability \cite{36}.

\par
Groundwater quality monitoring poses unique challenges compared to surface water monitoring due to aquifer’s positions and the difficulty in accessing underground water sources \cite{37}. Unlike surface water, which is readily available for sampling, groundwater exists within porous rock formations and soil layers, making direct observation and sampling more challenging \cite{38, 39}. The variability in aquifer depth, flow patterns, and geological composition further complicates monitoring efforts \cite{40, 41}. Since groundwater moves slowly through underground formations, contaminants may take years to spread or be detected, delaying pollution identification and remediation \cite{42}. Additionally, groundwater contamination is often invisible and can persist for long periods before its effects become noticeable, unlike surface water pollution, which can be more easily observed and addressed \cite{43, 44}. The limited number of accessible wells and boreholes restricts the ability to conduct widespread monitoring, making it difficult to obtain comprehensive data on water quality \cite{45, 46}. Sampling and testing groundwater require specialized equipment and techniques, increasing costs and logistical complexities \cite{47}. Furthermore, contamination sources such as industrial waste, agricultural runoff, and leachate from landfills can infiltrate deep into aquifers, making cleanup efforts difficult and expensive \cite{48, 49}. Despite these challenges, regular monitoring is crucial to ensure groundwater remains a safe and sustainable resource for drinking, agriculture, and industry \cite{50, 51}. Advanced approaches, such as real-time monitoring and machine learning models, offer promising solutions to enhance groundwater quality assessment and management \cite{52, 53}.

\par
Assessing groundwater quality requires consideration of both global and local factors, as various elements impact water safety and sustainability \cite{54}. On a global scale, climate change affects precipitation patterns and can lead to increased salinity in coastal groundwater due to saltwater intrusion \cite{55, 56}. Population growth places higher demand on water resources, leading to over-extraction and depletion \cite{57, 58}. Industrial waste and agricultural practices contribute to contamination through hazardous chemicals, nitrates, and pesticides \cite{59}. International organizations, such as the WHO, establish guidelines to ensure water quality standards (WHO, 2017) \cite{60}. Locally, geological formations and aquifer characteristics determine the natural mineral composition and vulnerability to contamination, while land use activities like urbanization and waste disposal directly impact water quality \cite{61, 62}. The reliance on groundwater for drinking and agriculture varies by region, making continuous monitoring and data collection essential for detecting contamination trends \cite{63}. Effective regulatory frameworks play a crucial role in maintaining sustainable groundwater extraction and pollution control \cite{64, 65}. By addressing both global challenges and regional conditions, comprehensive strategies can be implemented to safeguard and manage groundwater resources efficiently \cite{66}. Advancements in remote sensing and hydrogeological modeling offer new opportunities for more accurate groundwater quality assessments \cite{67}.

\section{Materials and Methodology}
\subsection{Study Area}
Odisha is a coastal state in the eastern part of India having over 41 million inhabitants. It shares its boundaries with the Bay of Bengal in east as portrayed in Figure 1. It is also among the eleventh largest states in the union in terms of population and the eighth largest state in terms of area. It is situated in the latitudes 17.780N and 22.730N, and in the longitudes 81.37E and 87.53E. It has an area of 155,707$km^2$, which is 4.87$\%$ of total area of India, and a coastline of 450 km. According to a Forest Survey of India report, it has 48,903$km^2$ of forest cover which is 31.41$\%$ of the state's total geographical area. The state is in the tropical region having humid climate, hot summers, good monsoon rainfalls and mild winters. The region is rich in alluvial deposits and the availability of minerals such as chromite, bauxite, coal and iron ore are greatly beneficial to the economic environment of the state \cite{68}. Groundwater is one of the most important natural resources in Odisha and is used for agricultural, drinking, and industrial purposes. However, the quality of groundwater is different from one area to another. Some areas of Odisha experience water pollution due to high levels of industrial or mining activities that lead to contamination of water by heavy metals like Zinc (Zn), Chromium (Cr(VI)), Iron (Fe), Copper (Cu), Cobalt (Co), Nickel (Ni), Sodium (Na), Potassium (K), Fluoride (F), Chloride (Cl). Such an example is Sukinda Valley in Jajpur district of Odisha, which is known to be India’s most polluted area. This valley has over 97$\%$ of India’s chromite reserves, making the quality of groundwater highly contaminated causing adverse human health and making the water not portable for drinking purposes \cite{69}. The main reason for the groundwater contamination of Sukinda Valley is due to the presence of hexavalent chromium ($Cr^{6+}$) which is a very toxic heavy metal. A report by the Indian Bureau of Mines (IBM) revealed that $Cr^{6+}$ levels have exceeded the permissible limit of 0.05mg/L which results in serious kidney failures in most of the local communities \cite{70}. In addition to that, many people are also suffering from lung infections, dermatitis, and liver damage as well. Also, the heavy mining and industrial activities have degraded the agricultural land present there, affecting the crop productivity. 

\begin{figure}[h!]
\begin{center}
\shadowbox{\includegraphics[width=16cm, height=10.8cm]{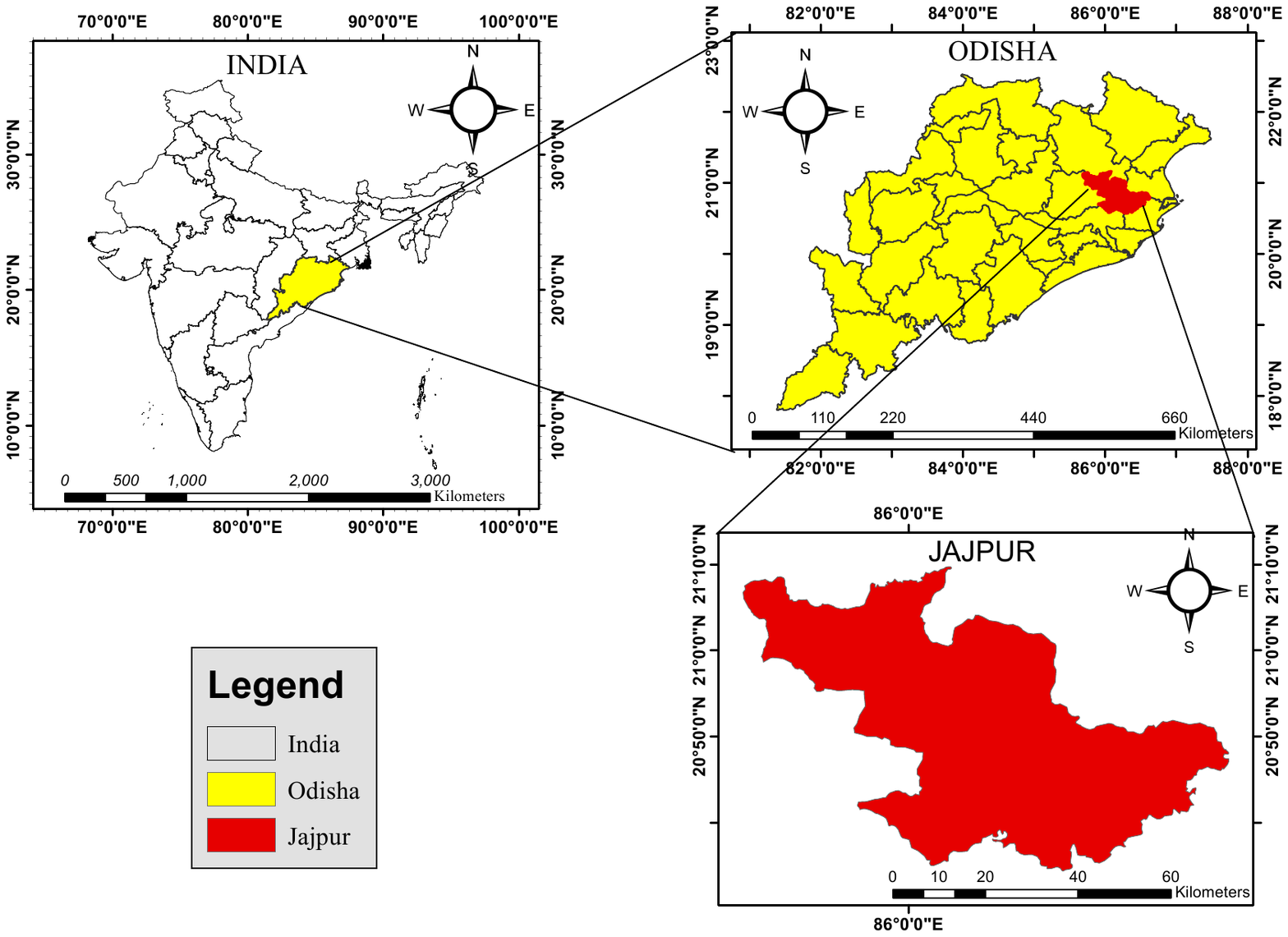}}
\caption{Location Map: Country Map showing the national context of the study area, State Map showing the regional location within the country, District Map demonstrating the area where the Sukinda Valley is located.  }
        \label{fig1}
\end{center}
\end{figure}

\subsection{Data Collection and Data Description}
For this study, dataset is obtained from the Central Ground Water Board (CGWB) under the Ministry of Jal Shakti, Department of Water Resources, River Development, and Ganga Rejuvenation, Government of India. For this study, data from 2019-2022 has been taken. This dataset consists of features like Well$\_$id, S.no, State, District, Taluka, Site, Latitude, Longitude, Year, pH, EC (Electrical Conductivity), CO3 (Carbon Carbonate), HCO3 (Bicarbonate), Cl (Chloride), SO4 (Sulphate),  NO3(Nitrate), PO4 (Phosphate), TH (Total Hardness), Ca (Calcium), Mg (Magnesium), Na (Sodium), K (Potassium), F (Fluoride), SiO2 (Silicon Dioxide), Block and Location\cite{71,72,73}.

 The dataset is having about 1989 records, providing proper insights of the groundwater quality indicators present in several districts of Odisha. Here, pH has a range from 4.99 to 8.85, having an average value of 7.844 and a standard deviation of 0.389. The EC is having a variation from 35µm/cm to 3650µm/cm with an average value of 681.35µm/cm having a standard deviation of 471.70µm/cm. TH is ranging from 6.00mg/L to 1235.00mg/L averaging 22.059mg/L with a standard deviation of 144.709mg/L. The mean concentration of Ca, Mg, Na, K, F, Cl are 45.87mg/L, 26.92mg/L, 48.88mg/L, 10.90mg/L, 0.38mg/L and 77.41mg/L respectively. A detailed description of the statistics is given in the following table:
 \\[0.5cm]

\begin{table}[h]
\begin{center}
\begin{tabular}{cccccccccc}
\hline
\hline
\rowcolor{cyan!50}   & pH & EC & Th & Ca & Mg & Na & K & F & Cl \\
\hline
\hline
                   Count & 1989 & 1989 & 1989 & 1989 & 1989 & 1989 & 1989 & 1989 & 1989 \\[0.2cm]

                   Mean & 7.844 & 681.356 & 220.059 & 45.870 & 26.922 & 48.880 & 10.901 & 0.385 & 77.419\\[0.2cm]

                   Std & 0.389 & 471.700 & 144.709 & 28.911 & 22.488 & 53.802 & 22.542 & 0.393 & 81.407 \\[0.2cm]

                   Min & 4.990 & 35 & 6 & 1 & 0 & 1 & 0 & 0 & 2 \\[0.2cm]

                   25$\%$ & 7.660 & 350 & 113 & 26 & 11 & 17 & 1.600 & 0.140 & 25 \\[0.2cm]

                   50$\%$ & 7.900 & 570 & 192 & 40 & 22 & 32 & 3.600 & 0.270 & 51 \\[0.2cm]

                   75$\%$ & 8.100 & 890 & 291 & 59 & 38 & 60 & 10.100 & 0.480 & 98 \\[0.2cm]

                   Max & 8.850 & 3650 & 1235 & 263 & 196 & 708 & 345 & 4.680 & 739 \\[0.2cm]
\hline 
\hline                 
\end{tabular}
\caption{
Statistical data description table. 
}
\end{center}
\end{table}

\subsection{Calculating Ground Water Quality Index (GWQI) score}
For calculating GWQI, a WQI model given by \cite{74} is used. It provides a proper approach to access groundwater quality by considering various water quality indicators (i.e. pH, Electrical Conductivity $\&$ concentration of heavy metals like Calcium (Ca), Magnesium (Mg), Sodium (Na), Potassium(K), Fluorine(F)) . Based on the concentration of various contaminants the GWQI is computed. It follows the following procedure:

\subsubsection{Selecting the required Water Quality Indicators (WQI)}
From the dataset obtained by Central Ground Water Board (CGWB), this study utilized physio-chemical water quality indicators like pH, Electrical conductivity (EC) and Total hardness (TH). Also, heavy metals water quality indicators like Calcium (Ca), Magnesium (Mg), Sodium (Na), Potassium (K), Fluoride (F), Chloride (Cl) are also used for calculating GWQI score \cite{72}.

\subsubsection{Calculating Sub-Index (SI)}
It is one of the main steps of GWQI score calculation as it transforms raw water quality indicator values into a standardized scale ranging from 0 to 100. If SI is equals to 0 then the water quality is poor and if SI is equals to 100 then the water quality is good \cite{72}. The SI basically helps in identifying the contribution of each indicator to the overall GWQI score. It is calculated using eq 1.

\begin{equation}
SI \equiv
\begin{cases}
\dfrac{WQI_{value}}{std_{min}} \times 100, if WQI_{value} < std_{min}\\
\\
\dfrac{std_{max}}{WQI_{value}} \times 100, if WQI_{value} > std_{max}\\
\\
100, Otherwise.
\end{cases}
\end{equation}

where, $WQI_{value}$ is the value of the water quality indicator present in the dataset, $std_{min}$ is the lower acceptable level of the water quality indicator and $std_{max}$ is the upper acceptable level of the water quality indicator.

\begin{table}[h]
\begin{center}
\begin{tabular}{|p{4.5cm}|p{3cm}|p{2cm}|p{2cm}|}
\hline
\rowcolor{cyan} Water Quality Indicator  & Limits used & $Std_{min}$ & $Std_{max}$ \\
\hline
  pH & (6.5,8.5) & 6.5 & 8.5 \\   
   \hline
  EC &  1500 µm/cm & 1 & 1500 \\ 
  \hline
  TH &  300 mg/L & 1 & 300 \\ 
  \hline
  Ca &  75 mg/L & 1 & 75 \\ 
  \hline
  Mg &  50 mg/L & 1 & 50 \\
  \hline
  Na &  200 mg/L & 1 & 200 \\ 
  \hline
  K &  12 mg/L & 1 & 12 \\ 
  \hline
  F &  1.5 mg/L & 1 & 1.5 \\ 
  \hline
  Cl &  250 mg/L & 1 & 250 \\                   
\hline                  
\end{tabular}
\caption{
Standard limits defined by the World Health Organization (WHO)
}
\end{center}
\end{table}

Here, for pH only, a range of (6.5, 8.5) is defined and for all other indicators (EC, TH, Ca, Mg, Na, K, F, Cl), only a single value is given and these values are the maximum permissible limits ($std_{max}$). The minimum permissible limits ($std_{min}$) are assumed to be 1, since negative values are not allowed in concentration-based parameters. In table 2, a detailed description of ($std_{min}$) and ($std_{max}$) values defined by WHO. 

\subsubsection{GWQI Score Aggregation function}
The work of an Aggregation function is to convert sub-index results into a single value that shows the overall groundwater quality\cite{73}. It helps in making decisions while consuming the water as lower the GWQI score better the water quality whereas higher the GWQI score suggest contamination. It can be calculated using eq 2.

\begin{equation}
GWQI =\sqrt{\sum_{i=1}^{n}(SI_{i})^{^{2}}}
\end{equation}

where, $SI_{i}$ is the index of the $i^{th}$ water quality indicator and n is the total number of indicators considered. After calculating the GWQI score, the next goal is to look into the Table 3 and find out the groundwater quality status.

\begin{table}[h]
    \centering
    \begin{tabular}{|p{3cm}|p{5.5cm}|p{5cm}|}
    \hline
       \textbf{GWQIScore} & \textbf{Groundwater Quality Status} & \textbf{Explaination}\\
    \hline
        0-50 & \cellcolor{green!60}Excellent & Water is safe for drinking and requires no treatment.\\
    \hline    
        50-100 & \cellcolor{green!10}Good & Water is slightly contaminated but still can be used by basic filtration. \\
        \hline    
        100-200 & \cellcolor{yellow!80}Poor & Water is not considered to be ideal for drinking purposes and requires proper treatment. \\
        \hline    
        200-300 & \cellcolor{orange!90}Very Poor & Water is highly contaminated and should be avoided for drinking purposes. \\
        \hline    
        300-400 & \cellcolor{red!90}Unsuitable for Drinking & Water is severely polluted and can be very hazardous to health.\\
    \hline    
    \end{tabular}
    \caption{Classification of Groundwater quality on the basis of GWQI score}
\end{table}

\subsection{Procedure for GWQI prediction}
In recent years, for predicting GWQI, ML has become a better choice for minimizing model uncertainty and making more reliable GWQI predictions. Researchers employ various ML algorithms to calculate the performance and accuracy of these predictive GWQI models \cite{74}. Figure 2 illustrates the workflow diagram which has been used in this study. A structured GWQI prediction technique using ML is as follows:

\subsubsection{Data pre-processing}
It is one of the most important steps before adding any data into the ML algorithms. It makes sure that data is in suitable format for further training procedures. The major parts of the data pre-processing process include data cleaning, feature selection, data standardization and data splitting.
\paragraph{Data Cleaning}
Data cleaning is considered to be the first and foremost step in data pre-processing. It ensures that the dataset is suitable for further analysis and is also consistent and accurate. This process involves various steps like filling null values, removing duplicates and outliers. Due to data entry errors or incomplete surveys many missing values may arise in the dataset. To handle these, the missing numerical data is filled using the mean by preserving the overall distribution and ensuring consistency. Likewise, the missing categorical data is filled using the mode hence maintaining the data integrity. Duplicate records occur in a dataset when the same data is present in more than one rows. This may happen due to accidental duplication in data entry, merging multiple datasets or due to some errors in data collection. It needs to be removed as it causes redundancy which results in more computational costs and increasing storage. Outliers are the data points that deviate from all other observations. As these extreme values can disturb the machine learning models, hence removing these outliers improves the model’s performance reducing the computational complexity. In this study the outliers are removed using the Interquartile Range (IQR) method. In this method, first we calculate the Interquartile Range (IQR) which is given by eq 3.

\begin{equation}
IQR = Q3-Q1	
\end{equation}

where, Q1 is the first quartile i.e. the value below which 25$\%$ of the data falls and Q3 is the third quartile i.e. the value below which 75$\%$ of the data falls. Hence, the middle 50$\%$ of the dataset is given by the Interquartile range (IQR). After calculating the IQR, the outlier bounds are calculated where the lower bound is given by Q1 $\times$ 1.5 and upper bound is given by Q3 $\times$ 1.5. Any data point which lies outside this outlier bound is considered as an outlier. After identifying the outliers, we filter those data points and visualize the results using boxplots.

\begin{figure}[h]
\begin{center}
\shadowbox{\includegraphics[width=17.5cm, height=9.8cm]{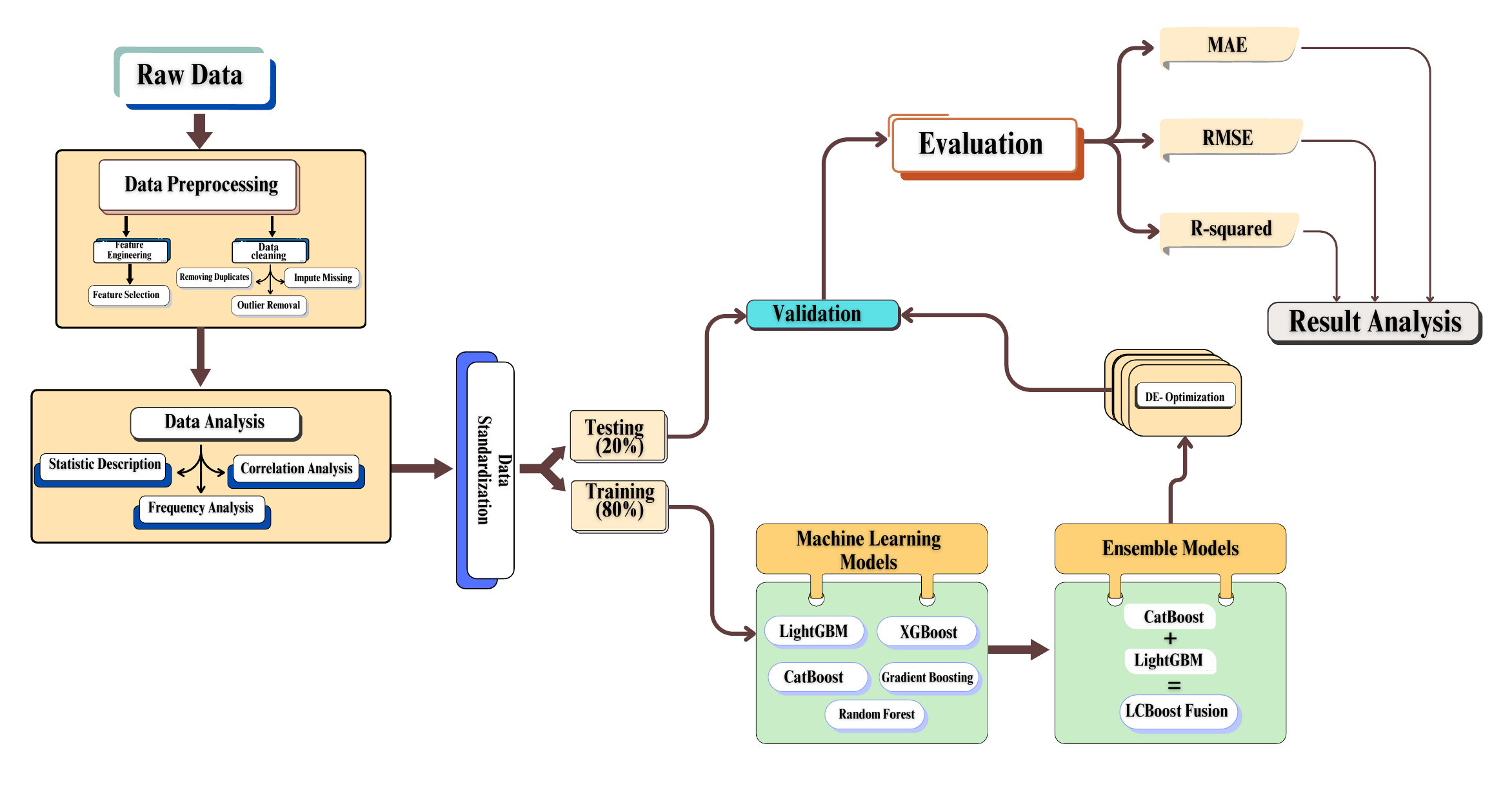}}
\caption{Workflow Diagram.  }
        \label{fig2}
\end{center}
\end{figure}

\paragraph{Feature Engineering}
In any ML problem, selecting the required features is considered to be one of the main steps. By removing all the irrelevant features helps in improving our model’s accuracy and making it more efficient. In this study from among all the features we selected nine most crucial groundwater quality indicators (pH, EC (Electrical conductivity), TH (Total hardness), Ca (Calcium), Mg (Magnesium), Na (Sodium), K (Potassium), F (Fluoride), Cl (Chloride)). These nine indicators were chosen because they together represent a combination of both physio-chemical and heavy metal water quality indicators. For our model, these nine combined features are chosen to be the independent variable and the calculated GWQI score has been chosen as the dependent variable (or target variable).

\paragraph{Data Standardization}
Data Standardization (also known as Data Normalization) helps in bringing all the groundwater quality indicators into a common scale. By doing so it reduces the number of model errors as all the groundwater quality indicators are in a uniform range \cite{75}. For standardization of the groundwater quality indicators, this study has used a feature scaling method in scikit-learn known as the StandardScaler which standardizes features by making the Mean($\mu$) = 0 and Standard deviation ($\sigma$) = 1. This method helps ML models work better by ensuring that all the indicators contribute equally to predictions. The StandardScaler method works by using the Z-score normalization formula which is given by eq 4.

\begin{equation}
Z = \frac{X-\mu}{\sigma}
\end{equation}

where, X is the original water quality indicator value, $\mu$ is the mean of the water quality indicator and $\sigma$ is the standard deviation of the water quality indicator.

\paragraph{Data splitting}
After scaling the data, it needs to be split into training sets and testing sets. The training set is used to train the ML model whereas the testing set is used to check the ML model’s performance. There are various ways of splitting the data like 80$\%$ and 20$\%$, 70$\%$ and 30$\%$, 60$\%$ and 40$\%$ or 50$\%$ and 50$\%$ \cite{75}. In this study we have used 80$\%$ of the data for training the model (i.e. learning the patterns in the data) and 20$\%$ of the data for testing the model (i.e. for evaluating the model’s accuracy). After splitting five ML algorithms are used for training and testing the dataset.

\subsection{Machine Learning Algorithms }
In this study Machine Learning models have been used for Regression analysis to model the relationship between independent variables and the dependent variable (or target variable). However, traditional regression methods such as linear regression, logistic regression or ridge regression often fail to capture nonlinear dependencies in data resulting to inaccurate predictions. However, advanced machine learning algorithms, especially ensemble learning methods, have predominantly improved the predictions by combining multiple weak classifiers into a strong model for predictions. Among these, gradient boosting techniques such as XGBoost, LightGBM, and CatBoost, along with conventional Gradient Boosting and Random Forest, have been used for comparing and evaluating the performance for predicted Groundwater Quality Index (GWQI). A brief overview of all the algorithms is given below:

\subsubsection{XGBoost (Extreme Gradient Boosting)}
Extreme Gradient Boosting or in short XGBoost, is an improved Gradient Boosting algorithm designed for high-performance regression and classification tasks. In addition to conventional Gradient Boosting, Extreme Gradient Boosting integrates advanced optimization techniques such as regularization (L1 and L2 penalties), tree pruning and parallel computation to improve the performance. This method has gained widespread acceptance due to its robustness against overfitting \cite{76}. XGBoost aims to solve the following optimization problem:

\begin{equation}
Objective (Obj) = \sum_{i=1}^{n} I(z_{i}, \hat{z}_{i})+\sum_{t=1}^{T} \Omega (f_{t})
\end{equation}

where, $I(z_{i}, \hat{z}_{i})$ is the loss function, $f_{t}$ = $W_{i}$ are the decision trees used in boosting, $\Omega(f_{t})$ is a regularization function that controls the complexity of individual trees, hence reducing the risk of overfitting. This function is defined as:

\begin{equation}
\Omega(f_{t}) = \gamma T + \frac{1}{2} \lambda \sum_{j=1}^{T} W_{j}^2
\end{equation}

where, T is the number of leaves in the tree, $W_{j}$  are the leaf weights and $\lambda$ and $\gamma$ are regularization parameters.

\subsubsection{LightGBM (Light Gradient Boosting Machine)}
Light Gradient Boosting Machine also known as LightGBM is another gradient boosting algorithm which is used to improve the accuracy of the predictions. The key idea for developing LightGBM was to reduce memory usage and training time by keeping the model’s performance intact. It was able to achieve this by using histogram-based learning, which involves binning data into histograms for faster computation. Also, instead of growing depth-wise, the leaves in this algorithm grows leaf-wise making it more efficient for large datasets \cite{77}. LightGBM also optimizes the same objective function as XGBoost:

\begin{equation}
Objective (Obj) = \sum_{k=1}^{n} I(z_{k}, \hat{z}_{k})+\sum_{p=1}^{T} \Omega (f_{p})
\end{equation}

where, $I(z_{k}, \hat{z}_{k})$ is the loss function, $f_{p}$ = $W_{n}$ are the decision trees used in boosting, $\Omega(f_{p})$ is a regularization function that controls the complexity of individual trees, hence reducing the risk of overfitting. This function is defined as:

\begin{equation}
\Omega(f_{p}) = \gamma T + \frac{1}{2} \lambda \sum_{n=1}^{T} W_{n}^2
\end{equation}

where, T is the number of leaves in the tree, $W_{n}$ are the leaf weights and $\lambda$ and $\gamma$ are regularization parameters.

However, LightGBM has two major additions:

\begin{itemize}
\item[1.] Gradient-Based One-Side Sampling (GOSS): In this method, data points having high gradient values are kept as representative and the rest are ignored. 
\item[2.] Exclusive Feature Bundling (EFB): In this approach, features that are not comorbid are combined which reduces the dimensionality of the dataset without much loss of information.
\end{itemize}

These innovations make LightGBM particularly suitable for dealing with large datasets.

\subsubsection{CatBoost (Categorical Boosting)}
Categorical Boosting or CatBoost is another gradient boosting algorithm which is used specifically to deal with categorical data efficiently. This is because XGBoost and LightGBM require a lot of pre-processing such as one hot encoding, but CatBoost encodes categorical features without reducing the information. It uses symmetric trees where splits are made for all the nodes based on the same feature \cite{78}. CatBoost minimizes the loss function:

\begin{equation}
L= \sum_{i=1}^{n} I(z_{i}, F(z_{i}))+\sum_{t=1}^{T} \Omega(f_{t})
\end{equation}

where, $F(z_{i})$ is the final prediction as an ensemble of weak learners, $I(z_{i}, F(z_{i}))$ is the error function and $\Omega(f_{t})$  is the regularization term to prevent overfitting.

However, CatBoost has two key advancements:

\begin{itemize}
\item[1.]Ordered Boosting: It is a technique to prevent target leakage through correct gradient estimation.  
\item[2.] Efficient Categorical Encoding: It reduces computational complexity with a specially designed encoding method.
\end{itemize}

\subsubsection{Gradient Boosting Regression (GBR)}
Gradient Boosting Regressor in short GBR sticks to a sequential learning process such that each new weak learner is trained with the help of the residual errors of the previous models. This method eventually gives better predictions by minimizing the loss function through gradient descent \cite{79}. Mathematically, the model is updated iteratively as follows:

\begin{equation}
F_{m+1}(x) = F_{m} (x) + \gamma ~h_{m}(x)
\end{equation}

where, $F_{m} (x)$ is the current model, $h_{m}(x)$ is the new weak learner trained on residuals and $\gamma$ is the learning rate.
The GBR model is able to capture complex relationships but is computationally intensive as compared to XGBoost and LightGBM.

\subsubsection{Random Forest Regressor}
Random Forest or random decision forests is an ensemble learning technique which was created for applications involving regression and classification tasks. It builds multiple decision trees and the average of their predictions is used to increase the generalization. Each decision tree makes its own decision without depending upon the other trees. It differs from boosting algorithms as it uses bootstrap aggregation (bagging) to decrease variance rather than making sequential refinements of weak learners \cite{79}. The final prediction is calculated by using the following formula:

\begin{equation}
\hat{z}_{i} = \frac{1}{M} \sum_{n=1}^{M} f_{m} (x)
\end{equation}

where, M is the total number of trees and $f_{m}(x)$ is an individual tree’s prediction.
Random Forest is quite robust to overfitting and noise which makes it suitable for high dimensional data.

\subsubsection{LCBoost Fusion (LightGBM + CatBoost)}
LCBoost Fusion is a hybrid machine learning model that combines both LightGBM and CatBoost for maximum predictive accuracy and efficiency. LightGBM is very fast in computation as it uses histogram-based learning, coupled with a leaf-wise tree growth strategy, making it quite efficient. On the other hand, CatBoost uses ordered boosting to reduce overfitting and has better management of categorical features which helps in improving the stability of the model. Thus, the two powerful algorithms combined in LCBoost Fusion are able to utilize their respective strengths to produce a well-balanced, high performing regression model. The integration is optimized further by Differential Evolution Optimization to ensure that weights are adapted in an adaptive manner for better generalization and precision.

The LCBoost Fusion model combines predictions from both LightGBM and CatBoost using optimized weights. 
Given a dataset, D = {($x_{i}$,$y_{i}$)}  $1 \leq i \leq n $
where, $x_{i}$ represents the feature vectors and $y_{i}$ represents the target values.

The individual models learn functions $f_{LGB} (x)$ and $f_{cat} (x)$ such that:
\begin{center}
$\hat{z}_{LGB} = f_{LGB}(x)$ and $\hat{z}_{Cat} = f_{cat} (x)$ 
\end{center}
where, $\hat{z}_{LGB}$ and $\hat{z}_{Cat}$  are the predicted values from LightGBM and CatBoost, respectively.

Each model minimizes the following loss function:

\begin{equation}
L(z,\hat{z}_{i}) = \frac{1}{n} \sum_{i=1}^{n} I(z_{i}, \hat{z}_{i})
\end{equation}

where, $I(z_{i}, \hat{z}_{i})$ is a predefined loss function i.e. Mean Squared Error (MSE) for regression:

\begin{equation}
MSE = \frac{1}{n} \sum_{i=1}^{n} (z_{i}-\hat{z}_{i})^{2}
\end{equation}

\paragraph*{LCBoost Fusion Model: Weighted Ensemble Approach}
The final prediction of LCBoost Fusion is given by linearly combining the predictions of the two models:

\begin{equation}
\hat{z}_{LCBoost~Fusion} = W_{Cat} \hat{z}_{Cat} + W_{LGB} \hat{z}_{LGB} + b
\end{equation}

where, $W_{Cat}$ and $W_{LGB}$ are the optimized weights for CatBoost and LightGBM respectively and b is a bias term that adjusts for systematic prediction errors.

To find the optimal weights, we minimize the Root Mean Squared Error (RMSE) between the actual and predicted values:

\begin{equation}
RMSE = \sqrt{{\frac{1}{n}} \sum_{i=1}^{n} (z_{i} - \hat{z}_{LCBoost~Fusion})^{2}}
\end{equation}

This optimization problem is solved using Differential Evolution (DE) which is a global optimization algorithm that iteratively refines the weights and bias.

\paragraph*{Optimization Using Differential Evolution (DE) :}

The goal of the optimization is to find the optimal set of weights (w1 , w2 , b) that minimizes the RMSE:

\begin{equation}
min\sqrt{{\frac{1}{n}}\sum_{i=1}^{n} \left(z_{i} - ((W_{Cat}\hat{z}_{Cat}+ W_{LGB} \hat{z}_{LGB}+ b))^{2} \right) }
\end{equation}

Subject to the constraints   $0.4 \leq W_{Cat} \leq 1.0$, $0.4 \leq W_{LGB} \leq 1.0$, $-5 \leq b \leq 5$.

The Differential Evolution (DE) Algorithm optimizes these parameters iteratively by evaluating a range of possible solutions and then analyzing their performance and selecting the best ones based on the RMSE metric.

\paragraph*{Advantages of LCBoost Fusion}
The combination of LightGBM and CatBoost in LCBoost Fusion generates a highly efficient and accurate regression model. The integration of LightGBM's speed and efficiency with CatBoost's strong categorical data handling abilities produces improved predictive performance with maintained computational efficiency. The use of Differential Evolution Optimization provides adaptive, data driven weighting mechanism to balance the contribution of both models at runtime, thus reducing overfitting and increasing generalization. The hybrid approach consistently outperforms standalone models which makes LCBoost Fusion a robust solution for complex regression tasks when both accuracy and efficiency are paramount.\\[1.5cm]

    \begin{longtable}{|p{17cm}|}
    \hline
     \cellcolor{blue!30}  \textbf{Algorithm1: Differential Evolution (DE) Optimizer}   \\
    \hline
    \\[1cm]
Input: Objective function f(x), population size NP, crossover rate CR, scaling factor F, max iterations T
Output: Optimized parameter set X*. \\
\vspace{3mm}
1. Initialize population P with NP random candidate solutions.\\
2. Evaluate fitness of each candidate in P.\\
3. While stopping criterion is not met (e.g., max iterations T reached):\\
a. For each candidate i in P:\\
i. Select three distinct candidates (r1, r2, r3) randomly from P.\\
ii. Generate mutant vector V = r1 + F * (r2 - r3)\\
iii. Generate trial vector U by crossover:\\
  \hspace{2cm} For each dimension j:\\
\hspace{2.5cm}If rand (0,1)$<$ CR or j is a randomly chosen index:\\
\hspace{3cm} U[j] = V[j]\\
    \hspace{2.5cm} Else:\\
    \hspace{3cm} U[j] = P[i][j]\\
      iv. Evaluate fitness of U.\\
      v. If fitness(U) $<$ fitness(P[i]):\\
          \hspace{2cm}  P[i] = U (Replace with better solution)\\
4. Return best solution X* from P.\\[1cm]
\\
    \hline    
    \caption{Differential Evolution (DE) Optimization Algorithm}
\end{longtable}

    \begin{longtable}{|p{17cm}|}
    \hline
     \cellcolor{blue!30}  \textbf{Algorithm2: LCBoost Fusion}   \\
    \hline
    \\
Given dataset $(X, y)$ $\in$ $R^{n \times m}$ $\times$ $R^{n}$:\\
1. Cross-Validation:  \\
        \hspace{2cm}Split data using KFold(n\_splits=10, random\_state=42) 
            \vspace*{1mm}\\
2. For each fold (X\_{train}, X\_{val}, y\_{train}, y\_{val}):\\
      \hspace{2cm}a. Preprocess:\\
      \hspace{2.5cm} Impute $\mu$(X\_{train}) $\rightarrow$ missing values\\
     \hspace{2cm} Scale: z = (X - $\mu$)/$\sigma$ using StandardScaler\\
       b. Train base models:\\
              \hspace{1.5cm} CatBoost: \\
                \hspace{2.5cm} f\_cat(x) =0\\
               \hspace{3cm}  for t=1 to 300 do:\\
        \hspace{3.5cm}f\_cat(x) = f\_cat(x)+ $\eta$·T(x,$\theta$\_t) \\
        \hspace{3.5cm} Minimize: $\Sigma$ L(y\_{i},\_cat(x\_i))+$\omega$($\theta$\_t)\\
         \hspace{3cm} for each t\\
         \hspace{2.5cm}where $\eta$=0.03, T = decision trees,\\
               \hspace{1.5cm} LightGBM: \\
                         \hspace{2.5cm}  f\_lgb(z) = 0\\
                         \hspace{3cm}  for t=1 to 300 d0:\\
                 \hspace{3.5cm}   f\_lgb(z) = f\_lgb(z) + $\eta$·h\_t(z) \\
                  \hspace{3cm} for each z\\
                    \hspace{2.5cm}  with $\eta$=0.03, h\_t = weak learners\\
 c. Hybrid optimization: \\ 
  \hspace{2cm}  Find $W_1$, $W_2$, b $\in$ R s.t.: \\ 
 \hspace{3cm}min  $\sqrt{(1/n\_train \sum(y\_i - (W_1f\_cat(x\_i) + W_2f\_lgb(z\_i) + b))^2)  }$\\
 \hspace{3cm} s.t. $W_1$,$W_2$ $\in$ [0.4,1.0], b $\in$ [-5,5] \\ 
      \hspace{2cm} Using differential evolution (maxiter=50)\\
       d. Compute metrics $\forall$  model m $\in$ {cat, lgb, hybrid}:\\
         \hspace{2cm} $RMSE_{m}$ = $\sqrt{(1/n \sum(y - {\hat{y}}_m)^2)}$ \\ 
        \hspace{3cm}   $MAE_m$ = 1/n $\sum$ $|y - \hat{y}_m|$  \\
        \hspace{2cm}   $R^2_m$ = 1 - $(\sum(y - \hat{y}_m)^2)/(\sum(y - \Bar{y})^2)$
                                        
\vspace*{5mm}
3. Aggregate metrics: E[RMSE], E[MAE], E[$R^2$] over folds\\
Output: Performance tables with E[.] for train/val\\

\\
    \hline 
    \caption{LCBoost Fusion Algorithm }   
    \end{longtable}
    
\section{Results}

\subsection{Data Pre-processing (Data Cleaning, Feature Engineering)}
Data Pre-processing ensures that before adding any data into the ML algorithms the data should be in suitable format for further training procedures. Data cleaning is considered to be the first step and involves various steps like filling null values, removing duplicates and outliers. To handle these, the missing numerical data is filled using the mean by preserving the overall distribution and ensuring consistency. Likewise, the missing categorical data is filled using the mode hence maintaining the data integrity. Duplicate records are required to be removed as it causes redundancy which results in more computational costs and increasing storage.

Figure 3 shows a feature correlation matrix heatmap that gives a proper identification and visualization of the relationships between various water quality indicators and the Groundwater quality Index (GWQI). It is a triangular heatmap where the correlation values range from -1 to +1, such that negative values represent an inverse relationship and positive values indicate a direct relationship.

\begin{figure}[h]
\begin{center}
\shadowbox{\includegraphics[width=17cm, height=10.8cm]{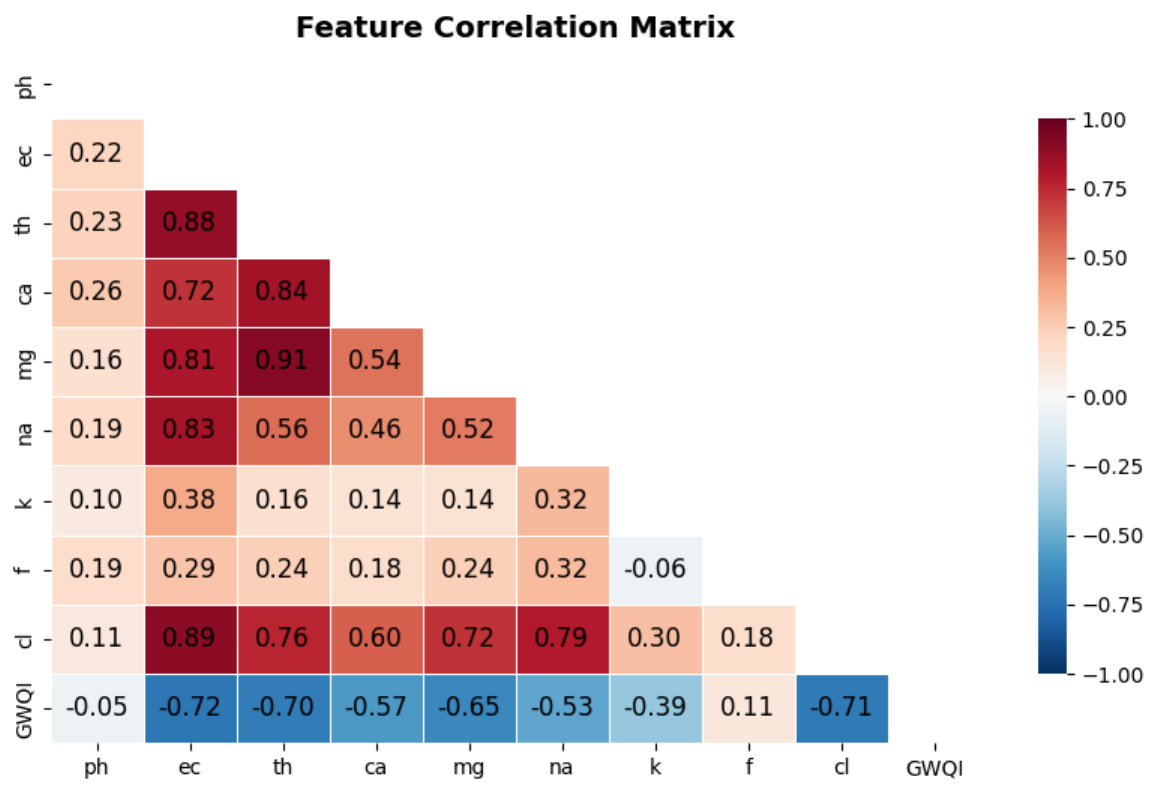}}
\caption{Feature Correlation Matrix.}
        \label{fig3}
\end{center}
\end{figure}

From the above figure for Strong Positive Correlation we can say that Electrical Conductivity (EC) has a high correlation with Chloride (Cl) (0.89), followed by Total Hardness (TH) (0.88), Magnesium (Mg) (0.81), Sodium (Na) (0.83). In the similar way, TH and Calcium (Ca) (0.84) as well as TH and Magnesium (Mg) (0.91) have strong correlations, which implies that with an increase in hardness-related features are closely associated with high levels of dissolved salts. However, the main motive of the correlation matrix is to highlight the water quality indicators that result in groundwater quality deterioration. This can be done by looking at the negative correlation between GWQI score and water quality indicators. From Figure 3, it is clear that, the water quality indicators like EC (-0.72), Cl (-0.71), TH (-0.70), and Na (-0.65), show strong negative correlations with the GWQI which further implies that higher concentrations of these water quality indicators adversely impact groundwater quality, leading to lower GWQI values. Outliers are the extreme values that deviate from all other observations which can further disturb the machine learning models. It can be detected with the help of a box plot as shown in Figure 4 which gives us a summary of the distribution of several water quality indicators in the dataset, highlighting the existence of outliers. The box plot shows the median, interquartile range (IQR), and extreme values for each water quality indicator, helping us to understand the probable anomalies and the variability in the dataset.

\begin{figure}[h]
\begin{center}
\shadowbox{\includegraphics[width=17cm, height=10.8cm]{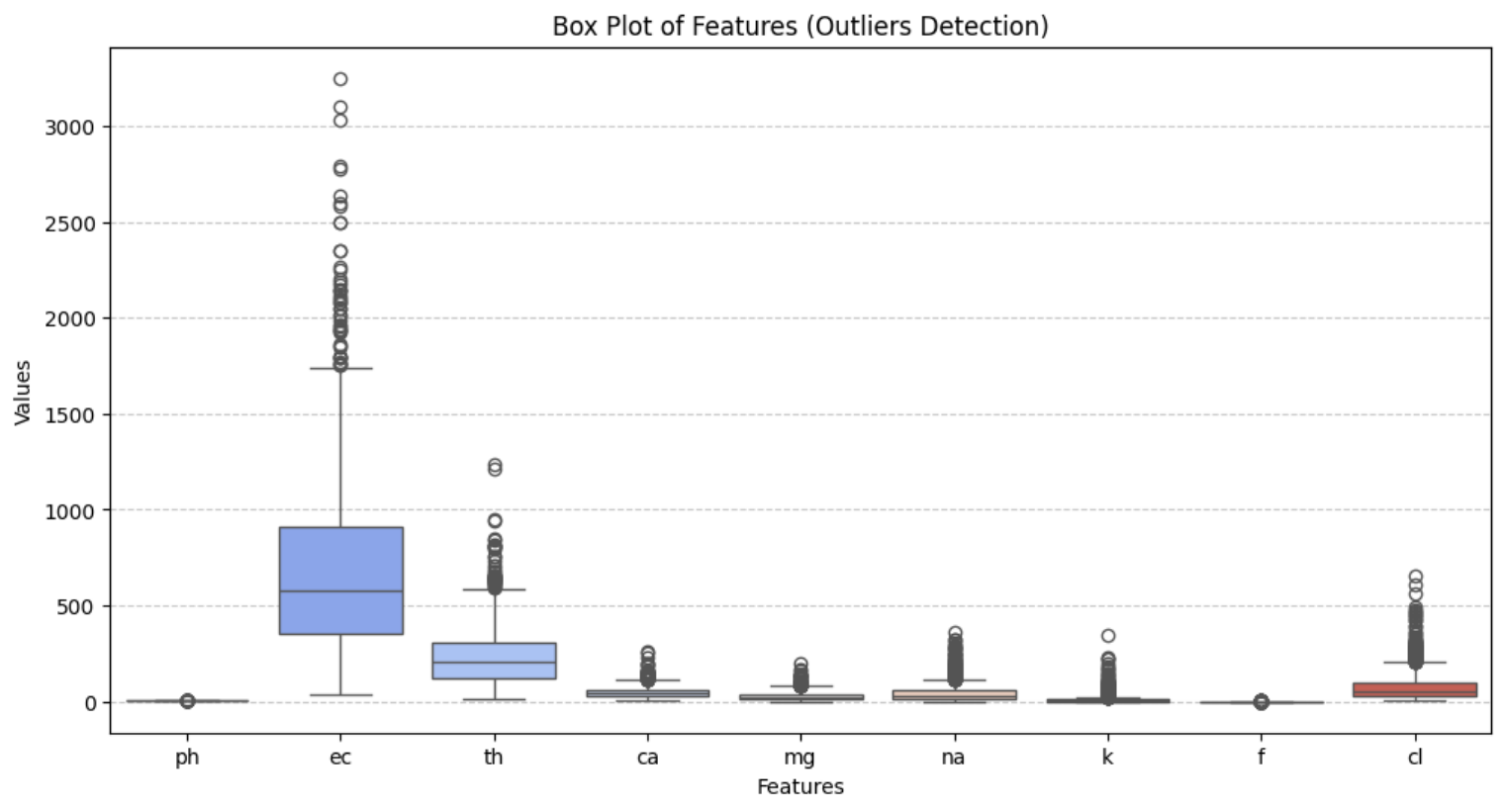}}
\caption{Box Plot of Features for Outlier Detection.}
        \label{fig4}
\end{center}
\end{figure}

From the above figure we can see that, EC has a large number of outliers, with values above 3000 $\mu$S/cm, showing very good water samples. Also, TH has shown some high values, indicating that certain water samples are very hard. Chloride (Cl) is also showing some wide distribution, which can be due to industrial discharge or agricultural runoff. Hence, the presence of such high-value outliers in EC, TH, and Cl implies potential groundwater contamination. Identifying and addressing these anomalies is crucial for groundwater management and ensures safe drinking water quality. This box plot analysis has served as an important analytical tool for data preprocessing, highlighting the need for further statistical treatments such as outlier removal before applying predictive models or classification techniques. The outliers in this study are removed using the Interquartile Range (IQR) method. In both Figure 5 and Figure 6 the left column of plots which are in red shows the original dataset, while the right column of plots which are in green shows the same features after applying the Interquartile Range (IQR) method to remove outliers. 

\par
In Figure 5 a comparative box plot analysis of various water quality parameters before and after outlier removal has been shown. From that figure we can see that after applying the IQR-based outlier removal technique, the data distribution becomes more compact, with a noticeable reduction in extreme values. The spread of values in the EC, TH, and Cl decreases, bringing them closer to a more normal distribution. In Figure 6 a comparative density plot analysis of various water quality parameters before and after the removal of outliers using the Interquartile Range (IQR) method.  Before outlier removal, several features such as EC, TH, and CL exhibited highly skewed distributions with extreme values. Post-outlier removal, the distribution of most features became more symmetric and continuous, indicating an improvement in data reliability. 

\par
After outlier removal, the data has been normalized using feature scaling method in scikit-learn i.e. StandardScaler to make sure that all water quality indicators have a mean(µ) of 0 and standard deviation of 1. Then, the dataset has been split into training and testing sets in the ratio 80:20. 

\begin{figure}
\centering
\begin{minipage}{.5\textwidth}
  \centering
\shadowbox{  \includegraphics[width=8cm, height=14cm]{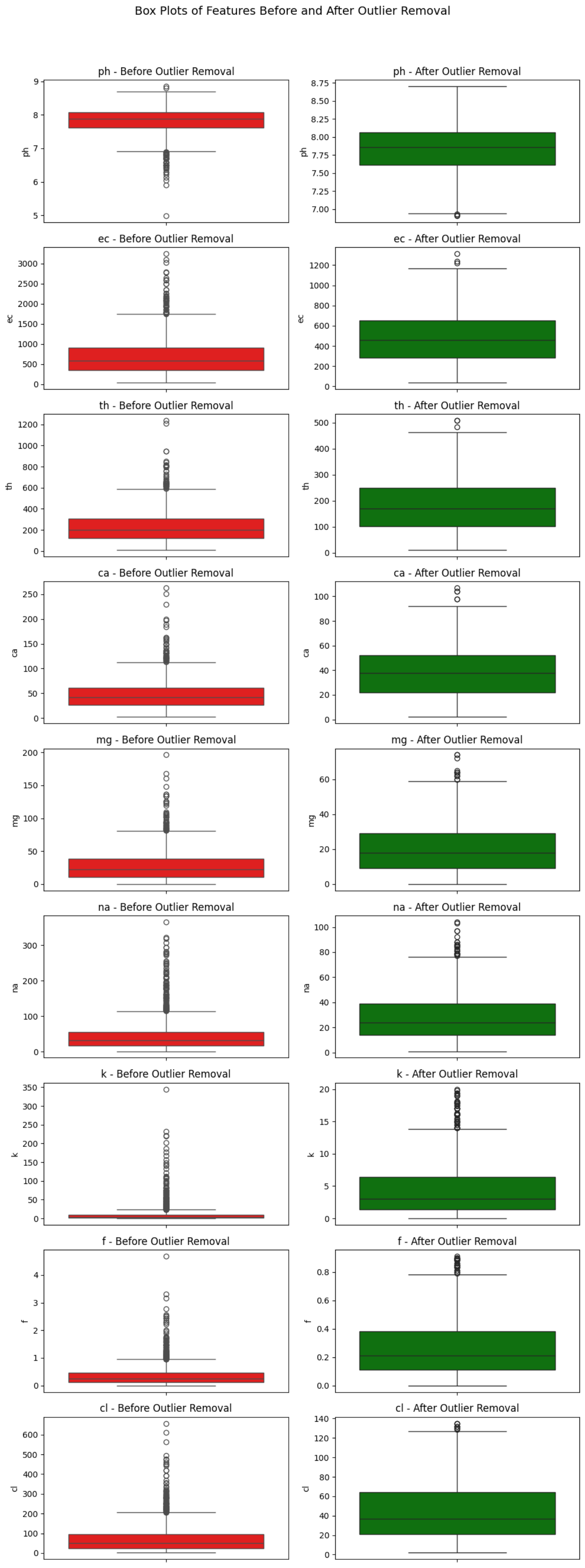}}
  \captionof{figure}{Boxplots of features before and \\ after Outlier Removal}
  \label{fig:test1}
\end{minipage}%
\begin{minipage}{.5\textwidth}
  \centering
\shadowbox{  \includegraphics[width=8cm, height= 14cm]{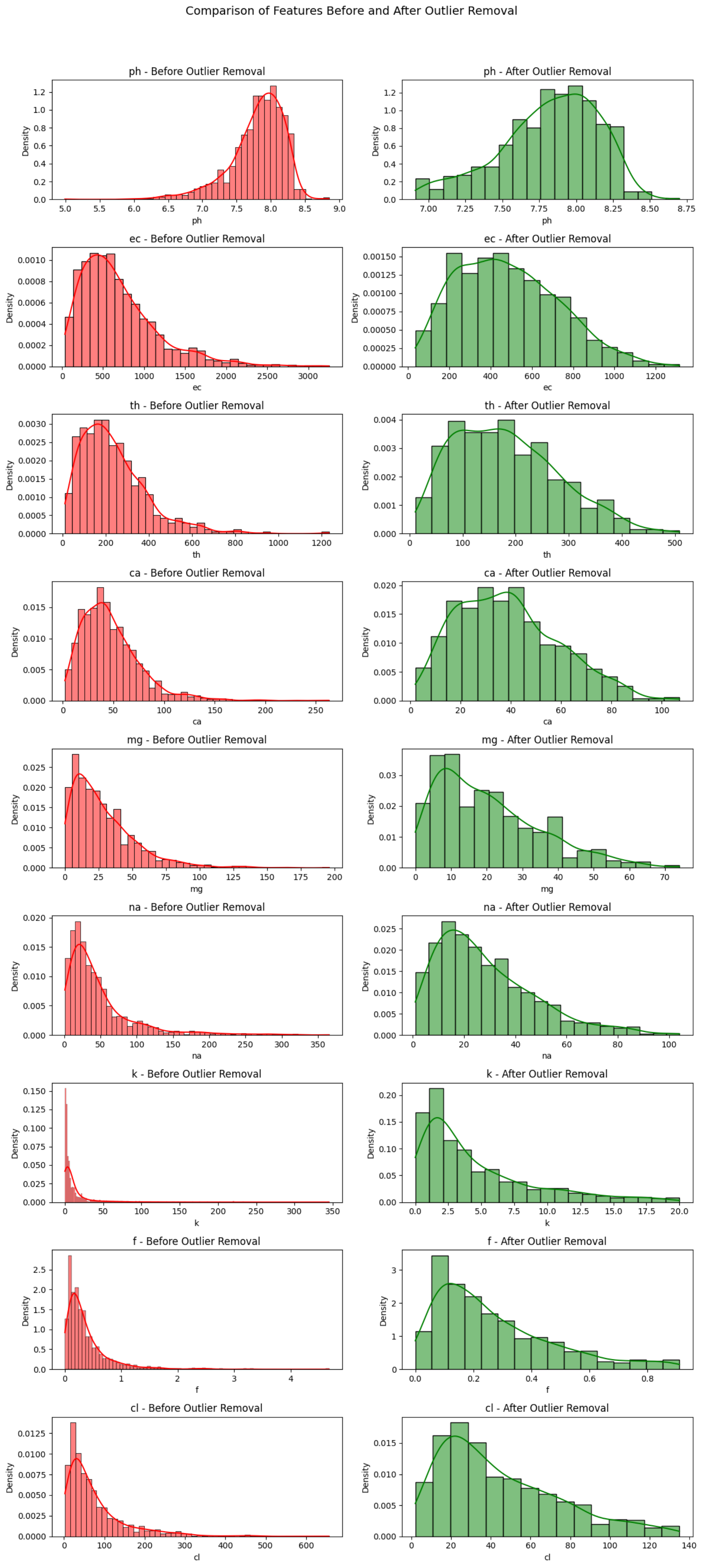}}
  \captionof{figure}{Comparison of features before and \\ after Outlier Removal using Histograms}
  \label{fig:test2}
\end{minipage}
\end{figure}

\subsection{Model Evaluation Metrics}
Model evaluation metrics are crucial for assessing the performance of machine learning models, mainly in regression tasks. Root Mean Squared Error (RMSE), Mean Squared Error (MSE), Mean Absolute Error (MAE), and R-squared ($R^2$) are mostly used to compare different models and determine their predictive accuracy. Equation 17 gives the Root Mean Square Error (RMSE) equation is a commonly used metric that measures the square root of the average squared differences between predicted and actual values \cite{80}. It is sensitive to large errors, making it useful for detecting outliers. The formula for RMSE is:

\begin{equation}
RMSE = \sqrt{\frac{1}{n} \sum_{i=1}^{n} (z_{i}-\hat{z}_{i})^{2}}
\end{equation}

A lower RMSE value indicates better predictive accuracy [83]. Equation 18 gives the Mean Square Error (MSE) equation that measures the average squared differences between predicted and actual values \cite{81}. It penalizes larger errors more than smaller ones, making it useful for identifying high variance in predictions. The formula is: 

\begin{equation}
MSE = \frac{1}{n} \sum_{i=1}^{n} (z_{i} - \hat{z}_{i})^{2}
\end{equation}

Lower MSE values indicate better model performance. Equation 19 gives the Mean Absolute Error (MAE) equation which calculates the average absolute differences between predicted and actual values . Unlike MSE, it treats all errors equally without squaring them. The formula is: 

\begin{equation}
MAE = \frac{1}{n} \sum_{i=1}^{n}\lvert{z_{i} - \hat{z}_{i}}\rvert
\end{equation}

Lower MAE values indicate more precise predictions. Equation 20 gives the R-Squared ($R^{2}$) Score that represents the proportion of variance in the dependent variable that is predictable from the independent variables\cite{82, 83}. It is defined as: 

\begin{equation}
R^{2} = 1 - \dfrac{\sum_{i=1}^{n}(z_{i} - \hat{z}_{i})^{2}}{\sum_{i=1}^{n}(z_{i} -\overline{z}_{i})^{2}}
\end{equation}

An $R^{2}$ value closer to 1 signifies a better fit, indicating that the model explains most of the variance in the target variable \cite{84}.

\subsection{Model Training Results}

\paragraph*{System Configurations:}

The computational experiments were carried out on a Windows 11 system with the following configuration: Windows version 10.0.26100, running on an x64-based architecture. The machine is equipped with a 12th Gen Intel (R) Core (TM) i5-12450H processor, featuring 8 cores and operating at a maximum clock speed of 2.0 GHz. The system has a total of 7.9 GB of physical memory (RAM). The storage system includes multiple drives: the C: drive has a total capacity of 156.6 GB, with 61.5 GB free, while the D: drive has 353.1 GB of total space, with 350.5 GB free. An additional G: drive is present with a total size of 161.0 GB, leaving 144.6 GB available for operations. The system is powered by Intel(R) UHD Graphics, and no dedicated NVIDIA GPU is detected, necessitating CPU-based computation for all tasks. This configuration ensures the system's capability to handle various workloads efficiently.

\begin{table}[h]
    \centering
    \begin{tabular}{|p{5cm}|p{2cm}|p{2cm}|p{2cm}|p{2cm}|}
    \hline
\rowcolor{blue!30}  Model     & RMSE  & MSE & MAE & $R^{2}$\\
    \hline  
          XGBoost & 0.1793 & 0.0321 & 0.0716 & 0.9643 \\
    \hline 
         \textbf{LightGBM}  &\textbf{ 0.1459} & \textbf{0.0213} & \textbf{0.0677} & \textbf{0.9763}\\
    \hline 
        \textbf{ CatBoost}  & \textbf{0.1148} & \textbf{0.0131} & \textbf{0.0572} & \textbf{0.9853}\\
    \hline 
          GradientBoosting (GB) & 0.1542 & 0.0237 & 0.0670 & 0.9736\\
    \hline   
          Random Forest    & 0.1829 & 0.0334 & 0.0773 & 0.9629\\
    \hline   
    \end{tabular}
    \caption{Evaluting Machine learning Models}
\end{table}

From the Table 6, we are able to analyze the performance of five machine learning models i.e XGBoost, LightGBM, CatBoost, Gradient Boosting (GB), and Random Forest by using the four model evaluation metrics i.e. Root Mean Squared Error (RMSE), Mean Squared Error (MSE), Mean Absolute Error (MAE) and $R^{2}$ Score ($R^2$). In Table 7 there is a detailed description of the hyperparameters used to predict the Groundwater Quality Index (GWQI) score during the training and evaluation process. To maximize the performance of the models CatBoost, LightGBM and LCBoost Fusion, the given hyperparameters have been decided to be used.

\begin{longtable}{p{5.1cm}p{5.3cm}p{5cm}}
\hline
\rowcolor{cyan!50} Model & Hyperparameter & Value  \\
\hline
CatBoost & n\_estimators & 300 \\
 
                 & learning\_rate
 & 0.03 \\

                 & max\_depth
 & 6 \\

                 & l2\_leaf\_reg
 & 3 \\

                 & random\_state
 & 42 \\        
                 
\hline
LightGBM & n\_estimators & 300 \\

                 & learning\_rate
 & 0.03 \\

                 & max\_depth
 & 4 \\

                 & num\_leaves
 & 31 \\

                 & Subsample
 & 0.7 \\

                 & colsample\_bytree
 & 0.8 \\
 
 \hline
 
LCBoost Fusion  & Weight\_CatBoost ($W_1$) & Optimized (0.4 - 1.0) \\

                 & Weight\_LightGBM ($W_2$)
 & Optimized (0.4 - 1.0) \\

                 & Bias (b)
 & Optimized (-5 to 5) \\

\hline
\caption{Hyperparameters selected in machine learning models}
\end{longtable}

\par 
From these models, CatBoost shows the best overall performance with the lowest RMSE (0.1148), MSE (0.0131), and MAE (0.0572), with highest $R^{2}$ score of 0.9853, showing superior predictive accuracy. LightGBM also performs very well, with an $R^{2}$ score of 0.9763 and RMSE (0.1459), MSE (0.0213), MAE (0.0677). We chose these models as they belong to the family of gradient boosting algorithms, which are convenient for structured data and provide high accuracy while handling complex patterns. XGBoost, LightGBM, and CatBoost are widely used for their efficiency and ability to mitigate overfitting, whereas Random Forest serves as a benchmark due to its robustness and interpretability. By observing the strong individual performance of CatBoost and LightGBM, we have proposed their fusion into a hybrid model i.e. LCBoost Fusion, to leverage their respective strengths. CatBoost efficiently handles categorical data and reduces overfitting, while LightGBM is optimized for speed and memory efficiency with large datasets. By integrating these models, LCBoost Fusion aims to further enhance generalization, reduce error rates, and provide a better predictive accuracy across diverse datasets.

\begin{table}[h]
    \centering
    \begin{tabular}{|p{5cm}|p{2cm}|p{2cm}|p{2cm}|p{2cm}|}
    \hline
\rowcolor{blue!30}  Model     & RMSE  & MSE & MAE & $R^{2}$ Score\\
    \hline 
         CatBoost  & 0.3534 & 0.1253 & 0.2271 & 0.9951\\
    \hline 
          LightGBM & 0.4309 & 0.1859 & 0.2128 & 0.9927\\
    \hline 
         \textbf{ LCBoost Fusion} & \textbf{0.3324} &\textbf{ 0.1106} &\textbf{ 0.1912} & \textbf{0.9956}\\
    \hline   
    \end{tabular}
    \caption{Performance of the model during training}
\end{table}

For GWQI prediction, LCBoost Fusion, which is the combined form of CatBoost and LightGBM models has been used to enhance the predictive performance. A 10-fold cross-validation strategy has been used to ensure better training and evaluation. The dataset underwent preprocessing, where missing values were imputed using column-wise means, and feature scaling was applied using StandardScaler to optimize model convergence, particularly for LightGBM. The CatBoost model was trained with 300 estimators, a learning rate of 0.03, and a maximum depth of 6, while the LightGBM model utilized an optimized hyperparameter values with a maximum depth of 4 and 31 leaves. To further refine predictions, a Differential Evolution (DE) optimization algorithm has been used to determine the optimal combined weights for LCBoost Fusion.From Table 8 we can clearly demonstrate that LCBoost Fusion outperformed both individual models, achieving the lowest training RMSE (0.3324), MSE (0.1106), and MAE (0.1912), with an exceptionally high $R^{2}$ score of 0.9956, signifying its superior ability to capture underlying patterns in the data.

\begin{table}[h]
    \centering
    \begin{tabular}{|p{5cm}|p{2cm}|p{2cm}|p{2cm}|p{2cm}|}
    \hline
\rowcolor{blue!30}  Model     & RMSE  & MSE & MAE & $R^{2}$ Score\\
    \hline 
         CatBoost  & 0.7507 & 0.6355 & 0.3448 & 0.9762\\
    \hline 
          LightGBM & 0.7605 & 0.6091 & 0.3569 & 0.9769\\
    \hline 
         \textbf{ LCBoost Fusion} & \textbf{0.6829} & \textbf{0.5102} & \textbf{0.3147} & \textbf{0.9809}\\
    \hline   
    \end{tabular}
    \caption{Performance of the model during validation}
\end{table}

For getting the generalization capability of the models, the same 10-fold cross-validation has been conducted on the unseen data. From Table 9 we can see that CatBoost model a validation RMSE (0.7507), MSE (0.6355), MAE (0.3448) and an R² score of 0.9763, while the LightGBM model also had a similar performance with an RMSE (0.7605), MSE (0.6091), MAE (0.3569) and an R² score of 0.9769. However, LCBoost Fusion has shown exceptionally better generalization performance, with lowest validation RMSE (0.6829), MSE (0.5102), MAE (0.3147) and the highest $R^{2}$ score of 0.9809, indicating reduced prediction error and improved predictive accuracy. The consistent outperformance of LCBoost Fusion across all validation metrics shows the effectiveness of ensemble learning and demonstrating that combining the strengths of multiple models can significantly enhance predictive accuracy. These findings establish LCBoost Fusion as a highly efficient and robust approach for predictive modeling, making it a valuable tool for complex real-world applications.

\paragraph*{LCBoost Fusion’s Edge over other Predictive Models}
Figure 7 shows the performance of three predictive models i.e CatBoost, LightGBM, and the LCBoost Fusion Model by comparing their predicted values against actual values. The red dashed line represents the ideal fit, where perfect predictions would lie. The CatBoost model demonstrates strong predictive accuracy, with most predictions closely following the ideal fit, though slight deviations are observed for lower actual values. LightGBM also performs well but exhibits greater dispersion in predictions, particularly in lower-value ranges, indicating higher variability.

\begin{figure}[h]
\begin{center}
\shadowbox{\includegraphics[width=17cm, height=5.8cm]{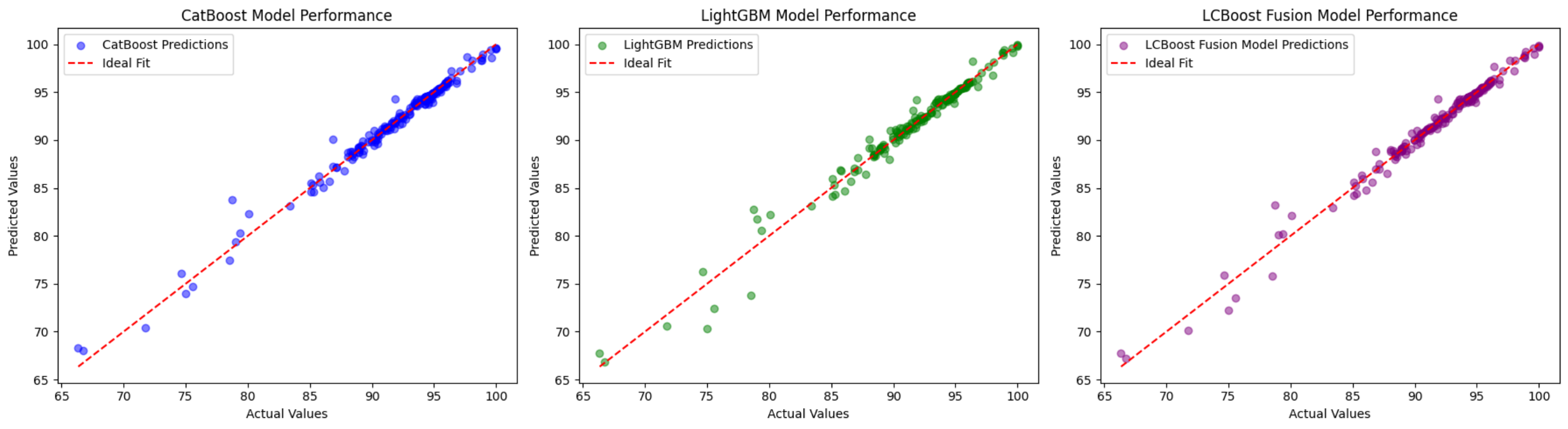}}
\caption{Performance Comparison of CatBoost, LightGBM and LCBoost Fusion.}
        \label{fig7}
\end{center}
\end{figure}

In contrast, the LCBoost Fusion Model shows the highest alignment with the ideal fit, effectively combining the strengths of both CatBoost and LightGBM. This suggests that the hybrid model reduces prediction errors and enhances generalization. Overall, the LCBoost Fusion Model outperforms the individual models, making it a promising approach for improving predictive accuracy in this context.

\paragraph*{Feature Importance Analysis for GWQI Prediction}
Figure 8 gives the feature importance analysis for Groundwater Quality Index (GWQI) prediction using the LCBoost Fusion Model, which integrates CatBoost and LightGBM. The bar chart ranks the features based on their contribution to the model’s predictive performance. Potassium (K) is identified as the most influential feature, contributing 34.15$\%$ to the model's predictions, followed by Fluoride (F) with 25.97$\%$ and Total Hardness (TH) with 22.10$\%$. Electrical Conductivity (EC) and Magnesium (Mg) also play significant roles, with respective importance values of 16.29$\%$ and 11.58$\%$. The remaining features, including Calcium (Ca), Chloride (Cl), Sodium (Na), and pH, exhibit comparatively lower importance, with pH having the least impact at 2.26$\%$.

\begin{figure}[h]
\begin{center}
\shadowbox{\includegraphics[width=17cm, height=10.5cm]{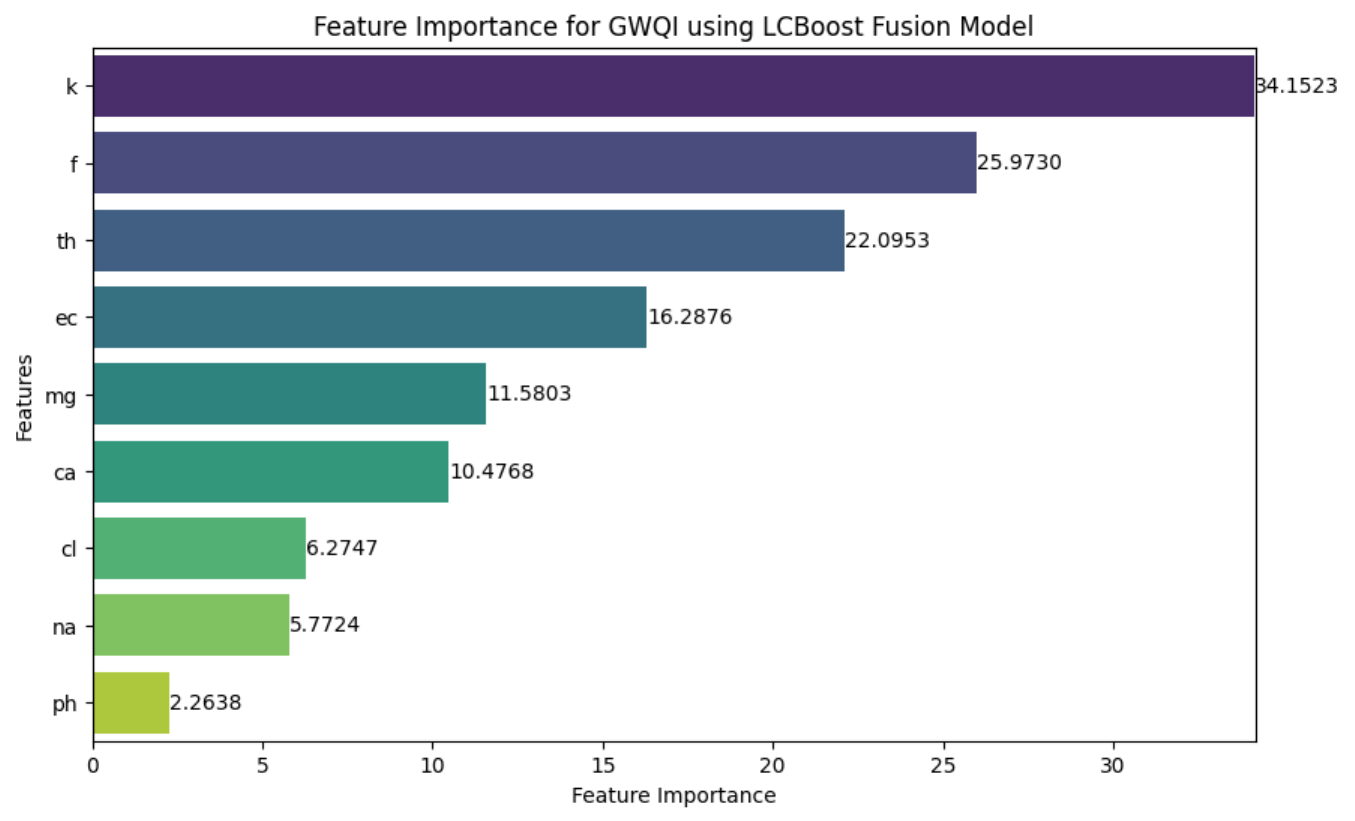}}
\caption{Feature Importance for Groundwater Quality Index Prediction.}
        \label{fig8}
\end{center}
\end{figure}

This analysis highlights that Potassium, Fluoride and Total hardness are the primary determinants of groundwater quality in the dataset. The lower importance of pH suggests that within the given data, pH variations contribute least to the GWQI prediction compared to other chemical constituents. The LCBoost Fusion Model’s weighted averaging approach effectively balances feature contributions from CatBoost and LightGBM, ensuring robust and well-calculated predictions. These findings offer valuable insights for policymakers and researchers, helping prioritize key water quality parameters for monitoring and management.

\section{Discussion}

\subsection{Summary}

This study presents a comprehensive Spatio - Temporal Groundwater Quality Risk assessment in Odisha using machine learning techniques. The research focuses on groundwater contamination in Sukinda Valley of Jajpur district due to chromite mining, leveraging a dataset from the Central Ground Water Board (CGWB) covering the years 2019-2022. The study employs a multi-step methodology, including data preprocessing methods (like feature engineering and data cleaning), and groundwater quality index (GWQI) prediction using advanced machine learning models such as CatBoost, LightGBM, and their combined model i.e. LCBoost Fusion. The results highlight the effectiveness of LCBoost Fusion in outperforming other models, demonstrating superior predictive accuracy with an RMSE of 0.6829, MSE of 0.5102, MAE of 0.3147, and $R^{2}$ score of 0.9809. Feature importance analysis identifies Potassium (K), Fluoride (F), and Total Hardness (TH) as the most influential factors affecting groundwater quality.

\subsection{Implications}
This study’s findings help the policymakers to establish specific region’s groundwater quality management guidelines ensuring a good health free from kidney and liver damage risks. The research provides a data-driven tools to assess the groundwater quality, which can help government agencies and environmental organizations for decision-making. Understanding and analyzing groundwater contamination levels is much required for protecting public health, as exposure to heavy metals may lead to severe health issues like kidney and liver damage. Moreover, the LCBoost Fusion model helps in transforming the ability of machine learning in figuring out environmental monitoring and mitigating the ecological risks.

\subsection{Limitations}
Although it has very promising results but the research has certain limitations like the dataset used in this study relies on data collected from Central Ground Water Board (CGWB), which may have inconsistent or missed records, thereby affecting the model’s predictive accuracy. Despite LCBoost Fusion model outstanding performance but its effectiveness in varying geographical regions with different climatic and geological conditions requires further validation. This research focused mainly on a subset of physio-chemical indicators and heavy metals. However. other pollutants, such as emerging pollutants (e.g., pharmaceuticals) or microbial contaminants have not been considered. 

\subsection{Future work}
Future research should focus on collecting data from a broader range of locations across Odisha so that it will help the improve model its predictive accuracy. Collaborating groundwater quality data with satellite-based remote sensing data will provide a more comprehensive sight of contamination shifts with spatial and temporal changes. Additionally, future research should also consider a broad range of pollutants like microbial pathogens, organic contaminants and pesticides so that a more better risk assessment analysis can be done. Also developing a mobile or web-based application for GWQI prediction will provide an edge to this research work as it will assist the stakeholders for making timely decision.

\paragraph*{Conclusion}
The application of machine learning techniques for predicting groundwater quality index (GWQI) with high accuracy has been successfully demonstrated in this study by combining advanced ML models such as CatBoost, LightGBM and introducing its hybrid as LCBoost Fusion. The results emphasized the importance of Potassium, Fluoride, and Total Hardness as the main indicators affecting groundwater quality. This study provides a robust method for predicting the Groundwater Quality Index (GWQI) score helping the environmental management and public health policymakers for monitoring and analyzing future risks. LCBoost Fusion, the hybrid model proposed in this study, has shown its outstanding performance over individual models with an RMSE of 0.6829, MSE of 0.5102, MAE of 0.3147, and an R² score of 0.9809. These results implies that LCBoost Fusion provides much enhanced predictive accuracy and reliability in predicting the GWQI score and analyzing the future risks at that area. This study highlights the ability of machine learning in handling the environmental challenges and emphasizing the need for further efforts in data-driven groundwater quality assessment to ensure safe and sustainable water resources for future generations.

\paragraph*{Declaration of competing interest}
The authors declare that they have no known competing financial interests or personal relationship that could have appeared to influence the work reported in this paper.

\paragraph*{Data availability}
Data will be made available on request.


\end{document}